\documentclass[runningheads]{llncs}
\usepackage[T1]{fontenc}
\usepackage{latexsym}
\usepackage{arydshln}
\usepackage{booktabs}
\usepackage{enumitem}
\usepackage{color}
\usepackage{graphicx}
\usepackage{booktabs}
\usepackage{amsmath}
\usepackage{amsfonts}
\usepackage{amssymb}
\usepackage{xcolor}
\usepackage{siunitx}
\usepackage{url}
\usepackage[misc]{ifsym}
\usepackage[ruled,linesnumbered]{algorithm2e}

\begin{document}
\title{Time Series Data Augmentation as an Imbalanced Learning Problem\thanks{This work was partially funded by projects AISym4Med (101095387) supported by Horizon Europe Cluster 1: Health, ConnectedHealth (n.º 46858), supported by Competitiveness and Internationalisation Operational Programme (POCI) and Lisbon Regional Operational Programme (LISBOA 2020), under the PORTUGAL 2020 Partnership Agreement, through the European Regional Development Fund (ERDF) and NextGenAI - Center for Responsible AI (2022-C05i0102-02), supported by IAPMEI, and also by FCT plurianual funding for 2020-2023 of LIACC (UIDB/00027/2020 UIDP/00027/2020)}}
%
%

\author{Vitor~Cerqueira\inst{1,2} \and
Nuno~Moniz\inst{4} \and
Ricardo~Inácio\inst{1} \and
Carlos Soares\inst{1,2,3}}
\authorrunning{V. Cerqueira et al.}

\institute{Faculdade de Engenharia da Universidade do Porto, Porto, Portugal \\
\email{vcerqueira@fe.up.pt}\and 
Laboratory for Artificial Intelligence and Computer Science (LIACC), Portugal \and
Fraunhofer Portugal AICOS, Portugal
\and
Computer Science Dept., Notre Dame University, Notre Dame, Indiana, USA
}
\maketitle              
\begin{abstract}
Recent state-of-the-art forecasting methods are trained on collections of time series. These methods, often referred to as global models, can capture common patterns in different time series to improve their generalization performance. However, they require large amounts of data that might not be readily available. Besides this, global models sometimes fail to capture relevant patterns unique to a particular time series. In these cases, data augmentation can be useful to increase the sample size of time series datasets.
The main contribution of this work is a novel method for generating univariate time series synthetic samples.
Our approach stems from the insight that the observations concerning a particular time series of interest represent only a small fraction of all observations. 
In this context, we frame the problem of training a forecasting model as an imbalanced learning task. 
Oversampling strategies are popular approaches used to deal with the imbalance problem in machine learning. We use these techniques to create synthetic time series observations and improve the accuracy of forecasting models.
We carried out experiments using 7 different databases that contain a total of 5502 univariate time series. We found that the proposed solution outperforms both a global and a local model, thus providing a better trade-off between these two approaches.

\end{abstract}

\section{Introduction}

Producing accurate forecasts is a valuable effort in various application domains, such as retail or economics. These forecasts help reduce uncertainty and enable better planning of operations \cite{kahn2003measure}. In many cases, the data consists of multiple univariate time series, such as the sales of a collection of retail products. 

Forecasting a group of time series can be done with a global forecasting method that learns a single model with all available time series. Global models can derive relevant patterns from different time series and improve their generalization performance. This approach contrasts with a local methodology, where a model is created for each time series in the database. Classical methods such as ARIMA \cite{hyndman2018forecasting} or exponential smoothing \cite{gardner1985exponential} follow a local approach. Machine learning algorithms, including deep neural networks or gradient boosting, tend to be trained on multiple time series \cite{bandara2020lstm}.

A key motivation for using a global model is the additional data. In general, larger training sets lead to better forecasting performance \cite{cerqueira2022case}. 
Still, a sufficiently large dataset might not be readily available even when using all available series. Besides this issue, global models sometimes fail to capture relevant patterns unique to a particular time series. In these cases, data augmentation can be useful to increase the sample size of time series datasets \cite{bandara2021improving}.

This work proposes a novel approach for synthetic data generation of univariate time series samples using a collection of time series. We hypothesize that global models can miss the nuances of a particular time series of interest due to an imbalance issue. This imbalance arises from the condition that the observations representing a time series of interest represent a small fraction of the whole dataset. For example, in a collection with 1000 equally-sized time series, the observations concerning a single one represent only 0.1\% of the total data points.
In this context, we frame the problem of training a forecasting model with several time series as an imbalanced learning task. Imbalanced domain learning is the field of machine learning devoted to learning from domains where the more important cases are rare and scarcely available on the data set. In our setting, the important cases are observations concerning a time series of interest.

Resampling strategies such as \texttt{SMOTE} \cite{chawla2002smote} are an effective approach to tackling imbalanced domain learning problems. These work by under- or oversampling the dataset to alter the data distribution.
In this work, we use oversampling methods to create synthetic data that improves the representativity of a particular time series in the training data. 
We propose the method Time Series Entity Resampler (\texttt{TSER}) that uses a resampling algorithm to create new observations for a particular time series (entity) of interest.
The goal of this method is to manage a better local-global trade-off for each entity, and thus improve the forecasting accuracy of models.

We carried out experiments using 7 datasets that contain a total of 5502 time series. We found that \texttt{TSER} performs better than standard global or local methods for the target time series, thus achieving a better trade-off between these two approaches.
In summary, the contributions of this work are the following:
\begin{itemize}
    \item A novel approach for building forecasting models using datasets involving multiple time series based on an imbalanced-domain learning perspective;

    \item \texttt{TSER}: a particular instance of this approach, which leverages oversampling strategies to augment time series datasets towards a particular time series of interest;

    \item An extensive set of experiments that validate the performance of \texttt{TSER} relative to state-of-the-art approaches.
\end{itemize}

The proposed approach \texttt{TSER} and the experiments reported in this work are available online\footnote{\url{https://github.com/vcerqueira/tser}}.

\section{Background}\label{sec:2}

This section provides a background to our work. 
We start by describing auto-regression approaches for time series forecasting based on univariate time series (Section \ref{sec:2.1}). Then, we explore how global methods leverage multiple time series to train a model (Section \ref{sec:2.2}). Finally, in Section \ref{sec:2.3}, we provide a brief introduction to imbalanced domain learning problems and explain how this topic relates to our work.

\subsection{Time Series Forecasting}\label{sec:2.1}

A univariate time series is defined as a time-ordered sequence of values $Y = \{y_1, y_2, \dots,$ $y_t \}$, where $y_i \in \mathbb{R}$ is the numeric value of $Y$ observed at time $i$ and $t$ is the length of $Y$.
We tackle time series forecasting tasks, where the goal is to predict the value of upcoming observations of the time series, $y_{t+1}, \ldots, y_{t+h}$, where $h$ denotes the forecasting horizon. 

Machine learning methods usually tackle this problem using an auto-regressive type of modeling. Each observation of the time series is modeled as a function of its past lags according to time delay embedding~\cite{bontempi2013machine}. 
Time delay embedding involves using an approach based on sliding windows to build a dataset $\mathcal{D}=\{x, y\}^t_{q+1}$ where $y_i$ represents the $i$-th observation and $x_i \in \mathbb{R}^q$ is the $i$-th corresponding set of $q$ lags: $x_i = \{y_{i-1}, y_{i-2}, \dots, y_{i-q} \}$. Accordingly, the objective is to train a model to learn the dependency $y_i = f(x_i)$. 

\subsection{Global Forecasting Models}\label{sec:2.2}

Forecasting problems often involve collections of time series. We defined a collection of $n$ time series as $\mathcal{Y} = \{Y_1, Y_2, \dots, Y_n\}$. 
Traditional approaches build a forecasting model for each time series within a collection. Such models are commonly known as local models \cite{januschowski2020criteria}.
Leveraging the data of all available time series can be valuable to build a forecasting model. The dynamics of the time series within a collection are often related, and a model may be able to learn useful patterns in some time series that have not revealed themselves in others. Models that are trained on collections of time series are referred to as global forecasting models \cite{godahewa2021ensembles}.

For global approaches, the time delay embedding framework described above, where each observation is modeled based on its lags, is applied to each time series in the collection.
Then, an auto-regression model is trained on the joint dataset that contains all transformed time series.
In effect, for a collection of time series the dataset $\mathcal{D}$ is composed of a concatenation of the individual datasets: $\mathcal{D} = \{\mathcal{D}_1, \dots,  \mathcal{D}_n\}$, where $\mathcal{D}_j$ is the dataset corresponding to the time series $Y_j$.

It is worth noting that, as pointed out by Hewamalage et al. \cite{hewamalage2021recurrent}, global forecasting approaches do not assume that the time series are dependent. That is, the lags of one series can be used to forecast the future values of another series. These techniques exploit information from multiple time series to estimate the parameters of the model. When forecasting the future of a time series, the main input to the global model is the recent past lags of that series.

The key motivation for a global forecasting approach is the additional data used for training. Machine learning algorithms tend to perform better with larger training sets \cite{cerqueira2022case}. 
However, sometimes there is insufficient data to train a global model. In other cases, a global model can miss the nuances of local components \cite{sousa2023intersecting}.
Data augmentation techniques can be useful to mitigate these issues \cite{bandara2021improving}. 
Another way to manage the trade-off between a global and a local model is with clustering \cite{sousa2023intersecting}. For example, Godahewa et al. \cite{godahewa2021ensembles}  clustering similar time series together and then train a global model by cluster.

\subsection{Imbalanced Domain Learning}\label{sec:2.3}

\subsubsection{Tackling class imbalance}

Imbalanced domain learning is the field of machine learning concerned with datasets with a skewed target distribution. In this type of dataset, most of the observations belong to one of the classes. Crucially, the minority class is often the most relevant. Imbalanced domain learning is a relevant topic in several domains, including fraud detection, or anomaly detection.

According to Branco et al. \cite{branco2016survey}, methods for handling imbalanced problems can be split into three categories: 
\begin{itemize}
    \item Data pre-processing: This type of solution involves changing the data distribution by transforming the dataset. This is typically carried out using resampling methods such as \texttt{SMOTE}~\cite{chawla2002smote} or \texttt{ADASYN}~\cite{he2008adasyn};

    \item Prediction post-processing: Instead of changing the input data, one can post-process the output of the model instead. For example, optimize the decision threshold of a probabilistic classifier;

    \item Special-purpose algorithms: Another approach is to tailor a particular learning algorithm to handle imbalanced datasets. For instance, incorporating a cost matrix in the training process of the model.
\end{itemize}

For a comprehensive read about strategies for handling imbalanced problems we refer to the excellent survey by Branco et al. \cite{branco2016survey}.
In this work, we will focus on resampling strategies (data pre-processing).

\subsubsection{Resampling strategies}\label{sec:resamplingmethods}

Some of the most popular solutions used to tackle class imbalance in classification problems are resampling methods. 
These approaches transform the training set to enhance the prevalence of the minority class. This involves a strategy based on undersampling the majority class, oversampling the minority class, or both. Since the resampling occurs before model fitting, these methods are agnostic to the learning algorithm.

The simplest resampling methods are random undersampling and random oversampling.
The first randomly selects instances from the majority class and removes them from the training set. The latter also selects instances at random but from the minority class. The selected instances are replicated in the training data.

\texttt{SMOTE} (Synthetic Minority Oversampling Technique) is a widely used oversampling approach. It works by synthesizing new examples based on existing ones. This is accomplished by interpolating between instances from the minority class within their neighborhood. More precisely, an observation from the minority class is selected at random, along with its $k$ nearest neighbors. Then, a new synthetic instance is created by interpolating between one of the $k$ neighbors and the selected observation. This process can be carried out several times, for example until the distribution of classes is balanced.
After the initial publication \cite{chawla2002smote}, several extensions of this method have been published (e.g. \cite{fernandez2018smote}).

\texttt{ADASYN} (Adaptive Synthetic) \cite{he2008adasyn} and Borderline SMOTE (\texttt{BSMOTE}) \cite{han2005borderline} are two other oversampling methods that follow a similar approach to \texttt{SMOTE} to creates new samples for the minority class. 
Both of these try to create samples near the decision boundary, which are, in principle, more difficult to learn. 

Informed undersampling of the majority class is also a common approach to deal with the imbalance problem that embeds some domain information in the selection of the majority class examples to be removed. One example of this approach is the Near-Miss method \cite{mani2003knn} (\texttt{NM}), which tries to retain only the instances from the majority class that are close to the decision boundary.

\section{Time Series Entity Resampling}\label{sec:3}

\begin{figure}[h]
\centering
\includegraphics[width=.95\textwidth]{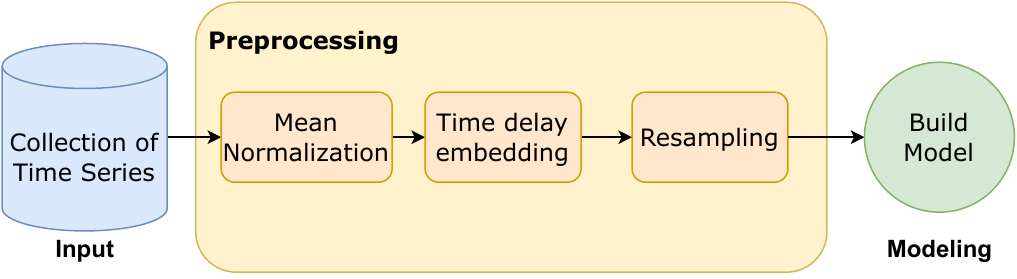}
\caption{Workflow behind \texttt{TSER}. The collection of time series is transformed for supervised learning using mean normalization and time delay embedding. New synthetic samples are created using oversampling. The resulting dataset is used to build a model.}
\label{fig:wf}
\end{figure}

This section formalizes the method \texttt{TSER}. 
As illustrated in Figure \ref{fig:wf}, the methodology fits within the preprocessing stage before building a forecasting model using a collection of time series.

\subsection{Data preparation}

Let $\mathcal{Y}$ denote a collection of univariate time series, where $Y_j$ denotes the $j$-th time series and $y_{j,i}$ represents the value of $Y_j$ at time $i$.

The set of time series in the collection can exhibit different value ranges. Therefore, we start by applying a mean normalization to bring all series into a common scale. This normalization can be defined as follows:

\begin{equation}
    y^{normalized}_{j,i} = \frac{y_{j,i}}{\frac{1}{t} \sum^{t}_{i=1} y_{j,i}}
\end{equation}

We denote the collection of normalized time series as $\mathcal{Y}^{normalized}$. 
Further preparation steps, such as logarithmic transformation for variance stabilization, can be applied as needed. For simplicity, we assume only mean scaling. Global forecasting methods and certain resampling methods involving distance calculations require a common value range to work optimally.
Henceforth, we will drop the $normalized$ superscript notation for conciseness. Unless stated otherwise, all subsequent steps are carried out using normalized time series.

The next step is to transform the collection of time series for supervised learning.  
This is done using the time delay embedding method based on a sliding window that is detailed in Section \ref{sec:2.1}. 
Using $h=1$ for illustration purposes, the goal is to  transform the collection of time series into a matrix structure such as the following:

\[
    \mathcal{D} = \left[
    \begin{array}{cccc|c}
    y_{1,1} & y_{1,2} & \dots  & y_{1, q} & y_{1, q+1} \\

    y_{1,2} & y_{1,3} & \dots  & y_{1, q+1} & y_{1, q+2}  \\

    \vdots & \vdots & \vdots  & \vdots & \vdots \\
    
    y_{j, i-p+1} & y_{j, i-p+2} & \dots & y_{j, i} & y_{j, i+1}\\
    
    \vdots & \vdots & \vdots  & \vdots &\vdots \\
    
    y_{n, t_n-p+1} & y_{n, t_n-p+2} & \dots   & y_{n, t_n} & y_{n,t_n+1}
    \end{array}
        \right]
    \]

\noindent where $t_n$ denotes the size of $Y_n$.
Each row in the matrix is a sample $($X$, $y$)$, where the last column denotes the target variable.

\subsection{Resampling}

Let $Y_k \in \mathcal{Y}$ denote a time series of interest within the collection.
We assume that either the size of the collection $\mathcal{Y}$ is small or $Y_k$ is under-represented in the dataset $\mathcal{D}$. In effect, we aim to create new synthetic time series samples concerning $Y_k$ to improve the modeling of this time series. We frame the problem as an imbalanced domain task and use resampling strategies (e.g., \texttt{SMOTE} \cite{chawla2002smote}) to tackle it.

To achieve this, we start by creating an auxiliary binary variable $b$. For a given sample $(x, y)$, $b$ is determined as follows:
\begin{equation}\label{eq:eq}
  b = \begin{cases}
            1 & \text{if } y \in Y_k ,\\
            0 & \text{otherwise}.
          \end{cases}
\end{equation}

\noindent Essentially $b$ takes the value of 1 if the respective sample comes from the time series of interest $Y_k$, or 0 otherwise. 
We use $b$ as an indicator variable to guide the creation of synthetic samples using resampling algorithms. These algorithms involve oversampling the minority class (time series of interest) to improve modeling.

For standard binary classification tasks, resampling strategies create new synthetic samples by interpolating the input variables with reference to the minority class. In this work, the minority class is a given time series of interest (when $b=1$). The input variables to the resampling algorithm are the concatenation of the input lags and the output future values ($x$, $y$), which represent a subsequence of a time series.
In effect, the resampling algorithm creates new ($x'$, $y'$) samples concerning the time series of interest ($b=1$). Since we preserve the input $x$ and output $y$ dependency, these samples have the same structure as the original data.

The created synthetic samples are represented in a dataset $\mathcal{D}'$, which is concatenated with the original dataset $\mathcal{D}$ to augment it. Subsequently, a forecasting model is trained based on the combined dataset $\mathcal{D} \cup \mathcal{D}'$.
This forecasting model has some global traits because the input data contains information from multiple time series. However, the model is on the local side of the local-global spectrum as each time series requires its own augmented dataset. This limitation is discussed further in Section \ref{sec:5}.

\section{Experiments}\label{sec:4}

This section presents the experiments used to validate the proposed method. These address the following research questions:

\begin{itemize}
    \item \textbf{RQ1}: Can we frame forecasting problems involving groups of univariate time series as an imbalanced learning problem and use resampling strategies to achieve a better trade-off between a local and global model?

    \item \textbf{RQ2}: What algorithm shows the best performance for resampling a dataset composed of groups of time series towards a particular entity?

    \item \textbf{RQ3}: What is the performance of each approach in other time series other than the time series of interest? 

    \item \textbf{RQ4}: How should the generated synthetic samples be combined with the original dataset?

    \item \textbf{RQ5}: How many synthetic samples should be created?
    
\end{itemize}

\subsection{Datasets}\label{ssec:data}

The experiments were carried out using 7 collections of univariate time series. A data summary is presented in Table \ref{tab:datasets}. The name of each data set refers to its identifier in the Python library \textit{gluonts}, where they were retrieved from.

\begin{table}[!bth]
	\centering
	\caption{Summary of the data sets}		
	\begin{tabular}{llrrr}
	\toprule
	\textbf{Name} & \textbf{Frequency} & \textbf{\# Time Series} & \textbf{Avg. Length} & \textbf{Horizon} \\
	\midrule
        
        \texttt{rideshare\_without\_missing} & Hourly & 2304 & 493 & 24 \\
        
        \midrule  
        
        \texttt{nn5\_daily\_without\_missing} & Daily & 111 & 735 & 14\\
        
        \midrule  
        
        \texttt{solar-energy} & Hourly & 137 & 7009 & 24\\
        
        \midrule  
        
        \texttt{traffic\_nips} & Hourly & 963 & 4001 & 24\\
        
        \midrule  
        
        \texttt{taxi\_30min} & Half-hourly & 1214 & 1488 & 48\\
        
        \midrule
        
        \texttt{m4\_hourly} & Hourly & 414 & 960 & 24\\
        
        \midrule
        
        \texttt{m4\_weekly} & Weekly & 359 & 934 & 12\\
        
	\bottomrule    
	\end{tabular}%
	\label{tab:datasets}
\end{table}

These datasets cover different application domains and sampling frequencies. Overall, the number of univariate time series is 5502.

\subsection{Experimental Setup}

\subsubsection{Data Preparation}

We fix the number of lags for time delay embedding to 10 in each problem.
For each dataset, the forecasting horizon is set according to the \textit{Horizon} column in Table \ref{tab:datasets}. For example, at each instance of a time series from the \textit{solar-energy}, the goal is to forecast the value of the next 24 observations. We adopt a direct approach for multi-step ahead forecasting \cite{bontempi2013machine}. This means that a model is built for each horizon. 

\subsubsection{Evaluation}

For each collection of time series, we apply a leave-one-time-series-out procedure for evaluating forecasting performance.
This means that each time series is iteratively picked for testing. In each iteration, the training set consists of the initial 70\% of observations of all time series (including the one used for evaluation). The test set comprises the last 30\% of observations of the time series in the respective iteration.

We use the mean absolute scaled error (MASE) as the evaluation metric. We also resort to the average rank and percentage difference between pairs of models to compare approaches across multiple time series.

\subsubsection{Methods}

We use the \texttt{lightgbm} regression algorithm to fit all models. This method has shown competitive forecasting performance, such as in the M5 forecasting competition \cite{makridakis2022m5} that also involved multiple time series.
We optimize the parameters of the \texttt{lightgbm} using 200 iterations of random search. The different configuration possibilities are detailed in Table \ref{tab:lgbm_parset}.

\begin{table}[!bth]
	\centering
	\caption{Pool of parameters used to optimize the \texttt{lightgbm}}		
	\resizebox{0.85\textwidth}{!}{%
	\begin{tabular}{lll}
    	\toprule
    	
    	\textbf{ID} & \textbf{Parameter} &  \textbf{Value}\\
    	    
	    \midrule   
        
        \texttt{num\_leaves} & Max \# of leaves per tree & \{3, 5, 10, 15\} \\
        
        \midrule   
        
        \texttt{max\_depth} & Max depth per tree & \{-1, 3, 5, 10, 15\} \\
        
        \midrule   
        
        \texttt{lambda\_l1} & L1 regularization & \{0.1, 1, 10, 100\} \\
        
        \midrule   
        
        \texttt{lambda\_l2} & L2 regularization & \{0.1, 1, 10, 100\} \\

	    \midrule   
        
        \texttt{learning\_rate} & Learning rate & \{0.05, 0.1, 0.2\} \\
        
        \midrule   
        
        \texttt{min\_child\_samples} & Min \# of points per leaf & \{7, 15, 30\} \\
        
        \midrule   
        
        \texttt{boosting\_type} & Base algorithm & gbdt \\
        
        \midrule   
        
        \texttt{num\_boost\_round} & Boosting iterations & 200 \\
        
        \midrule   
        
        \texttt{early\_stopping\_rounds} & Early stopping & 30 \\
        
		\bottomrule    
	\end{tabular}%
	}
	\label{tab:lgbm_parset}
\end{table}

We include two state-of-the-art methods that provide a reference performance:

\begin{itemize}
    \item \texttt{Local}: The traditional approach that fits a model using the historical observations of the time series of interest;
    \item \texttt{Global}: A global forecasting approach that fits a model using the historical observations of all available time series.
\end{itemize}

Then, we include four versions of \texttt{TSER} that vary in the resampling algorithm employed. Three of these are oversampling methods, namely \texttt{SMOTE}, \texttt{ADASYN}, and \texttt{BSMOTE}. While our focus is to create new samples using oversampling, we also include an undersampling approach (\texttt{NM}) in the experiments. We denote the application of \texttt{TSER} with these as \texttt{TSER}($m$), where $m$ is the respective resampling method (e.g. \texttt{TSER(SMOTE)}). The resampling methods used in the experiments are described in Section \ref{sec:2.3}.

Concerning the application of the resampling algorithms, we create (oversampling) or remove (undersampling) the number of samples necessary to make the dataset balanced. This means that the final dataset contains the same number of observations for both the time series of interest and the remaining time series. We will carry out an analysis to explore the sensitivity of the method to the sampling ratio.
All these algorithms apply a nearest-neighbor algorithm to drive the resampling process. In the experiments, we set the size of the neighborhood to 10. The remaining parameters of the resampling methods were left as default.

\subsection{Results}

\subsubsection{Average ranks and Bayesian analysis}

We start by analyzing the average rank of each method across all time series irrespective of the collection. A given method has a rank of 1 in a given problem if it shows the best performance (lowest MASE) in that problem. In effect, the average rank represents the average relative position of a method.

\begin{figure*}[hbt]
\centering
\includegraphics[width=.95\textwidth]{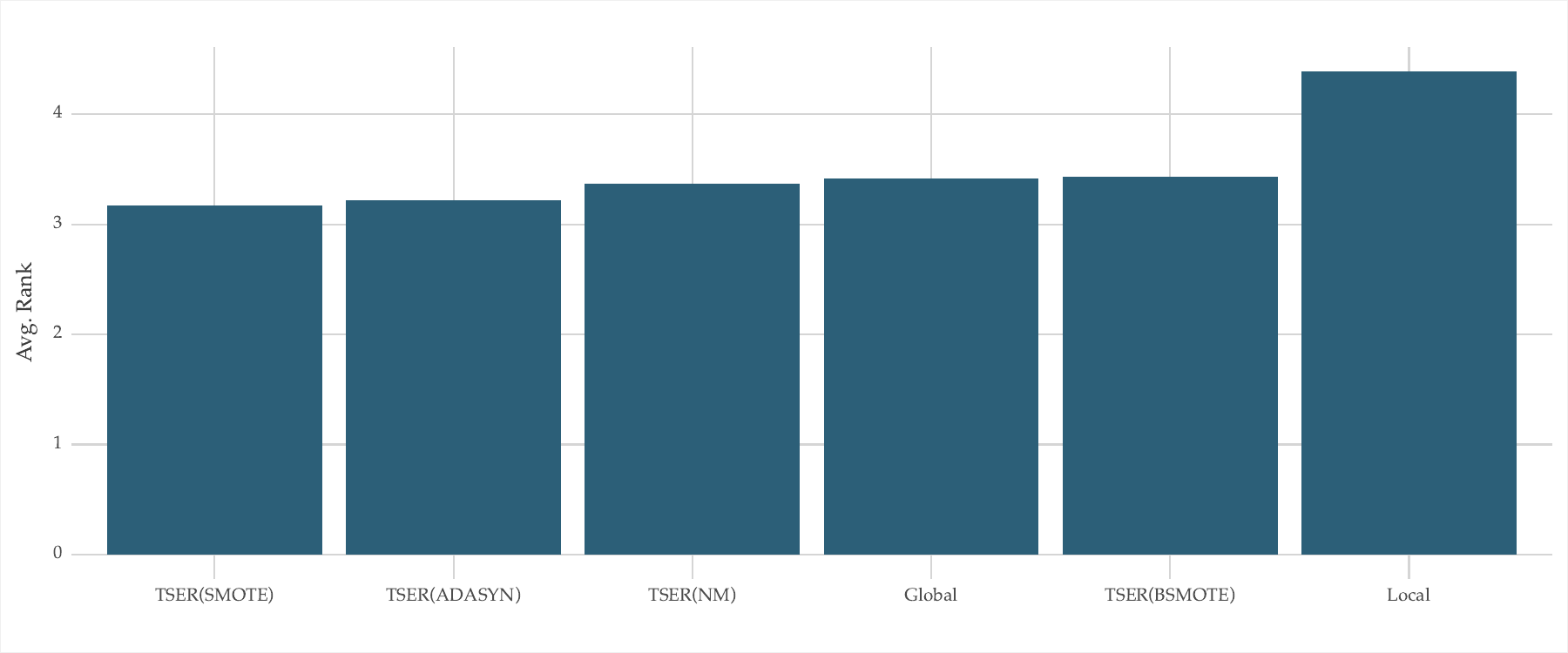}
\caption{Average rank of each method across all time series.}
\label{fig:avgrank1}
\end{figure*}

The results are presented in Figure \ref{fig:avgrank1}.
The top three positions are occupied by variants of the proposed method \texttt{TSER}, namely \texttt{TSER(SMOTE)}, \texttt{TSER(ADASYN)}, and \texttt{TSER(NM)}.

We break the average rank analysis by each collection of time series in Table \ref{tab:avg_rank_tab}.
\texttt{TSER(SMOTE)} shows the best or second-best score in 4 out of the 7 problems. The \texttt{Local} and \texttt{Global} reference methods show the best score on one problem each, though \texttt{Global} performs better overall.

As emphasized before, we focus on using oversampling algorithms to create new instances. Overall, the oversampling approaches used in the experiments perform better than \texttt{TSER(NM)} that applies undersampling with \texttt{NearMiss}. 
Yet, \texttt{TSER(NM)} shows the best average rank in 1 out of 7 problems (taxi\_30min).

\begin{table}
\caption{Average rank of each method in each dataset. Bold and underlined values represent the best and second-best scores.}
\label{tab:avg_rank_tab}
\resizebox{0.65\textwidth}{!}{%
\begin{tabular}{lrrrrrrr}
\toprule
 & \rotatebox{90}{m4\_hourly} & \rotatebox{90}{m4\_weekly} & \rotatebox{90}{nn5\_daily} & \rotatebox{90}{rideshare} & \rotatebox{90}{solar-energy} & \rotatebox{90}{taxi\_30min} & \rotatebox{90}{traffic\_nips} \\
\midrule
Global & 5.9 & \underline{2.6} & 3.7 & \textbf{2.8} & 2.9 & \underline{2.8} & 5.0 \\
Local & \textbf{1.8} & 5.1 & 5.6 & 4.8 & 5.0 & 3.9 & 4.5 \\
TSER(SMOTE) & 3.5 & \underline{2.6} & \textbf{2.1} & 3.3 & \textbf{1.8} & 3.9 & \textbf{2.2} \\
TSER(ADASYN) & 3.8 & \textbf{2.5} & 2.4 & 3.3 & \textbf{1.8} & 3.9 & 2.5 \\
TSER(BSMOTE) & 4.1 & 3.1 & \underline{2.3} & \underline{3.1} & 3.5 & 4.9 & \underline{2.4} \\
TSER(NM) & \underline{2.0} & 5.1 & 4.9 & 3.7 & 6.0 & \textbf{1.6} & 4.4 \\
\bottomrule
\end{tabular}%
}
\end{table}

The average rank analysis does not consider the magnitude of performance differences. We analyzed this aspect by computing the percentage difference of each method relative to \texttt{Global} and \texttt{Local}--the reference methods. For each time series, we compute the following value for each method: $$ 100 \times (\text{MASE}_{method} - \text{MASE}_{referece}) / \text{MASE}_{reference} $$

\begin{figure*}[hbt]
\centering
\includegraphics[width=.95\textwidth]{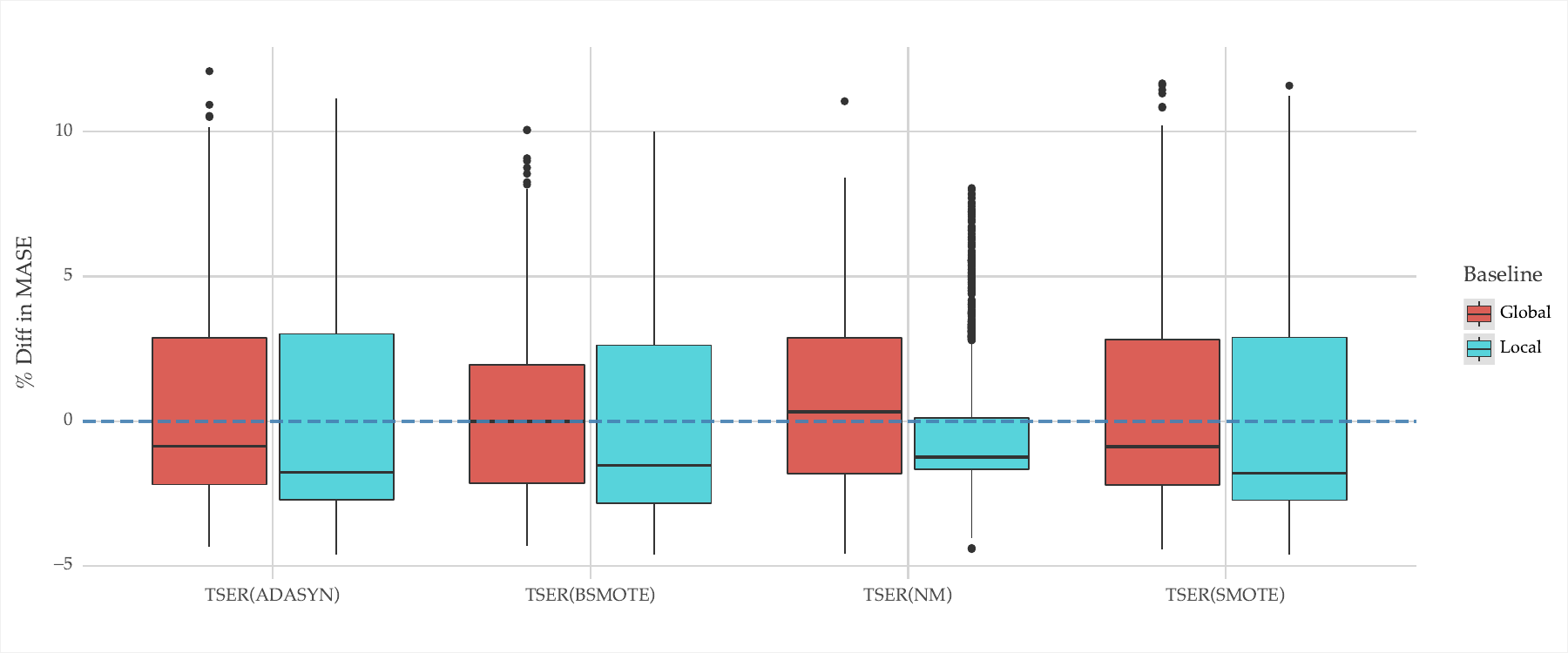}
\caption{Percentage difference in MASE between the respective method and each reference approach across all time series. Negative values denote better performance by the respective method.}
\label{fig:pd_baseline_all}
\end{figure*}

The results are shown in Figure \ref{fig:pd_baseline_all}. Overall, the results corroborate those obtained in the average rank analysis. Most \texttt{TSER} variants outperform both reference methods. The exceptions are \texttt{TSER(BSMOTE)} and \texttt{TSER(NM)} which perform comparably with \texttt{Global}.

A Bayesian analysis is carried out to assess the significance of the results using the Bayesian signed-rank test \cite{benavoli2017time}. We compare each approach with the  \texttt{Global} reference method.
The region of practical equivalence for the test is set to the interval [-5\%, 5\%]. This means that two methods are considered practically equivalent if their percentage difference in performance falls within this interval.
The results are shown in Table \ref{tab:bayes}, which describes the probability of each outcome (\texttt{Global} winning, \texttt{Global} losing, or a draw).

\begin{table}
\caption{Results of the Bayesian signed-rank test, which details the probability of each outcome.}
\label{tab:bayes}
\begin{tabular}{lrrr}
\toprule
 & \texttt{Global} wins & draw & \texttt{Global} loses \\
\midrule
\texttt{Local} & 1.00 & 0.00 & 0.00 \\
\texttt{TSER(SMOTE)} & 0.00 & 0.01 & 0.99 \\
\texttt{TSER(ADASYN)} & 0.00 & 0.21 & 0.79 \\
\texttt{TSER(BSMOTE)} & 0.00 & 1.00 & 0.00 \\
\texttt{TSER(NM)} & 1.00 & 0.00 & 0.00 \\
\bottomrule
\end{tabular}
\end{table}

The statistical test confirms the previous results. \texttt{TSER(SMOTE)} and \texttt{TSER(ADASYN)} have an high probability of outperforming \texttt{Global}, but the comparison with \texttt{TSER(BSMOTE)} ends up in a draw. The undersampling variation \texttt{TSER(NM)} loses against \texttt{Global}, which, in turn, wins against \texttt{Local}.

\subsubsection{Performance on Other Time Series}

The main limitation of \texttt{TSER} is that, by creating synthetic samples for a particular time series, the model will, in principle, perform worse in other time series.
We quantified this effect according to the difference between the average performance on the time series of interest and the average performance on other time series of the same collection. We focus this analysis on the \textit{nn5\_daily\_without\_missing} dataset for efficiency reasons.

\begin{figure*}[hbt]
\centering
\includegraphics[width=.95\textwidth]{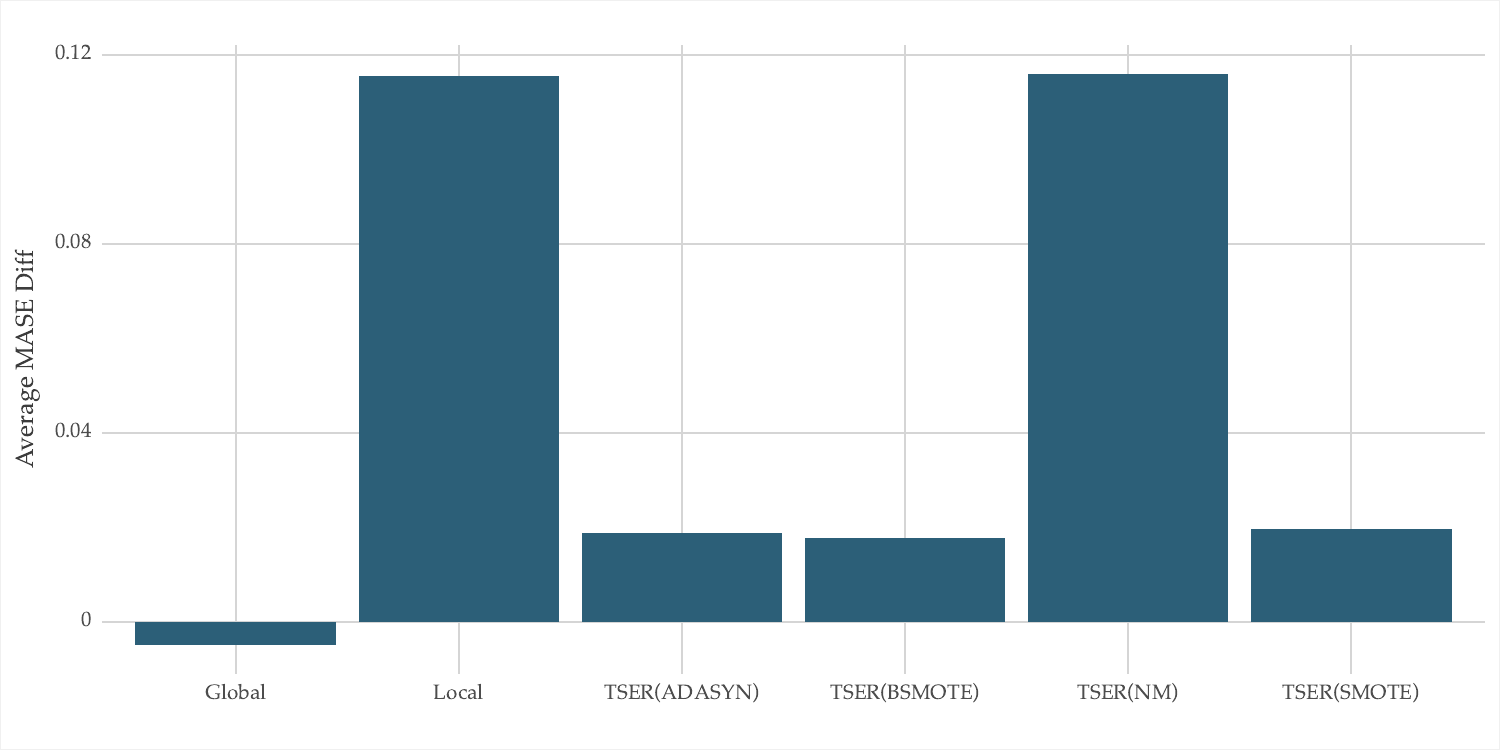}
\caption{Average difference in MASE between each approach when applied to the time series of interest and when it is applied to other time series of the same collection. Positive values denote a decrease in performance when the respective approach is applied in other time series.}
\label{fig:avg_pd_on_extra}
\end{figure*}

Figure \ref{fig:avg_pd_on_extra} shows the results of this analysis. For example, on average, the MASE of \texttt{Local} increases about 0.12 when the model is used in other series. 
Overall, the results of this analysis are expected. \texttt{Global}, since it is not tailored for any particular time series, shows a performance difference around zero. Conversely, \texttt{Local} is built using data from the time series of interest only. So, its performance decreases considerably when it is applied to other series. The same applied to \texttt{TSER(NM)}, which gives up data from other series to balance the dataset towards a particular time series.
While the oversampling versions of \texttt{TSER} also decrease their performance in other series, this effect is smaller than when undersampling or when using \texttt{Local}.

\subsubsection{Sensitivity analyses}

The results reported so far provide evidence about the benefits of using \texttt{TSER} to oversample a collection of time series towards a particular one.
In this section, we study the behavior of \texttt{TSER} to different configurations regarding the following aspects:
\begin{itemize}
    \item Data integration: how the generated synthetic samples should be combined with the original data;
    \item Sampling ratio: how many synthetic samples should be created
\end{itemize}

In the interest of conciseness, we focus on applying \texttt{TSER} with SMOTE for oversampling and on the \textit{nn5\_daily\_without\_missing} time series collection. 

\paragraph{Data integration}

The proposed application of \texttt{TSER} involves creating new synthetic samples for a particular time series of interest, and adding these to the original dataset that contains data from the whole collection. 
We studied alternative applications of \texttt{TSER} this aspect. Specifically, we compare the following variants:
\begin{itemize}
    \item \texttt{TSER}: The version of \texttt{TSER} that was used in the previous experiments, where an oversampling method is used for creating synthetic data for the time series of interest to balance the dataset

    \item \texttt{TSER(Local)}: A combination of \texttt{TSER} and \texttt{Local}, where besides oversampling the time series of interest to balance the dataset we then discard the data from other series

    \item \texttt{TSER(all)}: A variant of \texttt{TSER} in which we oversample the data of both classes (the class representing the time series of interest and the other class denoting the other time series). We first oversample the time series of interest to balance the dataset like in \texttt{TSER}. Then, we also oversample the data concerning other series in such a way that it increases its size by 50\%. 
\end{itemize}

The results are shown in Table \ref{tab:sens1} which presents the average rank of each approach, including the reference methods.

\begin{table}
\caption{Average rank and respective standard deviation of each approach}
\label{tab:sens1}
\resizebox{0.75\textwidth}{!}{%
\begin{tabular}{lllll}
\toprule
\texttt{Global} & \texttt{Local} & \texttt{TSER} & \texttt{TSER(Local)} & \texttt{TSER(all)} \\
\midrule
2.74$\pm$0.9 & 3.98$\pm$0.5 & \textbf{1.72$\pm$0.7} & 4.83$\pm$0.4 & \underline{1.74$\pm$0.6} \\
\bottomrule
\end{tabular}%
}
\end{table}

\texttt{TSER} and \texttt{TSER(all)} show a comparable score, which is better than the score of the remaining approaches. This suggests that creating synthetic samples for other series besides the target one does not affect the behavior of the proposed method.
The \texttt{TSER(Local)} score suggests that discarding the data from other series leads to poor performance. This means that, while the synthetic samples improve forecasting accuracy, the data concerning the other series is also valuable

\paragraph{Sampling ratio}

\begin{figure*}[th]
\centering
\includegraphics[width=.95\textwidth]{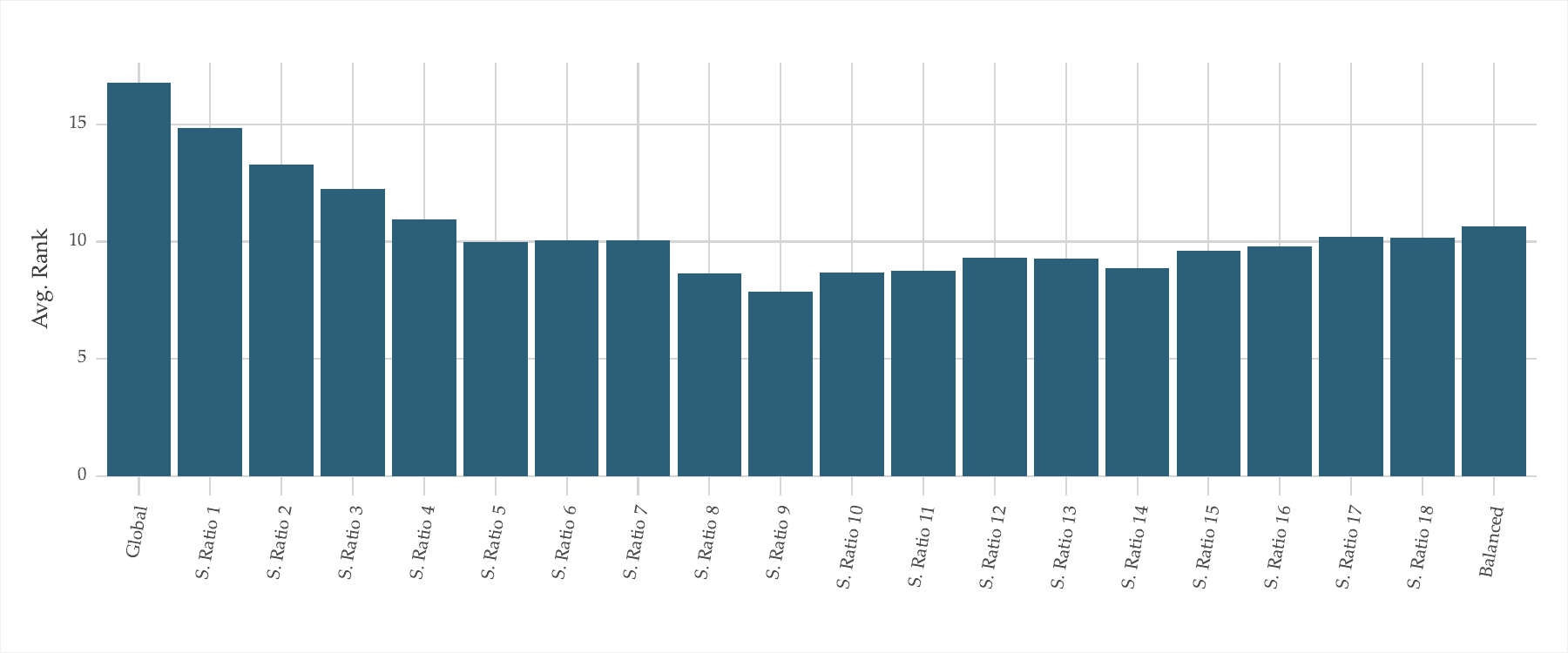}
\caption{Average rank of \texttt{Global} and \texttt{TSER} with varying sampling ratios.}
\label{fig:avgrank2}
\end{figure*}

Another relevant aspect concerning \texttt{TSER} is how many instances should be generated. In other words, the sampling ratio between the time series of interest and the other time series. 
So far, we applied \texttt{TSER} to balance the distribution. Suppose the dataset contains 100 observations for the time series of interest and 900 observations concerning others (a sampling ratio of 9:1). Then, \texttt{TSER} is used to create 800 instances and bring the sampling ratio to 1:1.
Balancing is often the default approach when using resampling methods, but it can be a suboptimal setting.

We analyze the sensitivity of \texttt{TSER} to different sampling ratios. We test 20 different ratios, starting from \texttt{Global} (no resampling, highest ratio imbalance) and ending in a balanced dataset (1:1 ratio). In between, we create  18 variants with uniformly distributed sampling ratios.
We denote these variants from 1 to 18, where 1 is the closest to \texttt{Global} and 18 is the closest to a 1:1 balance. Note that we did not use the actual ratio as notation since it can vary because of time series with varying length.

The results are shown in Figure \ref{fig:avgrank2}, which shows the average rank of each approach. In the figure, the methods are ordered by sampling ratio, with the leftmost approach being the most imbalanced one.
The average rank improves as we increase the size of the synthetic samples, but only up to a point where the effect stabilizes. 
The results suggest that the best sampling ratio is somewhere around 2:1.

\section{Discussion}\label{sec:5}

\subsection{Experimental results}

The experiments presented in the previous section show that using oversampling algorithms to create synthetic time series samples improves the forecasting accuracy on a time series of interest (\textbf{RQ1}).
Overall, applying \texttt{TSER} with \texttt{SMOTE} led to the best average rank score. A Bayesian analysis confirmed that the differences in performance are significant relative to state-of-the-art approaches (\textbf{RQ2}).

\texttt{TSER} is designed to improve the global-local trade-off on a particular time series. However, the experiments show that the approach decreases forecasting accuracy when it is applied to other time series besides the target one (\textbf{RQ3}). 

We carried out a sensitivity analysis concerning how to combine the generated synthetic samples with the original dataset. The results show that, while the new samples improve forecasting accuracy, the original data concerning other time series is also important (\textbf{RQ4}).

Finally, we also analyze how many samples should be created. New synthetic samples are beneficial only up to a certain size, after which the effect stabilizes. Overall, the best sampling ratio, estimated on a particular collection, is about 2:1. In such case, one-third of the samples of the final dataset represent the time series of interest (\textbf{RQ5}). 

\subsection{Limitations and future work}

The main limitation of \texttt{TSER} is that the generated synthetic observations are designed for a particular time series. Even though we rely on a collection of time series, the model trained on the augmented dataset is on the local side of the global-local spectrum. This means that, to get the most of \texttt{TSER}, we need to train a model for each time series in the collection. This represents a drawback concerning computational demands, which we plan to address in future work.

In this work, we focused on using \texttt{TSER} to create synthetic samples for a particular time series. This was done with oversampling methods such as \texttt{SMOTE} or \texttt{ADASYN}. However, we also tested an undersampling method, Near-Miss, that removed samples concerning the remaining time series. While oversampling provided overall better results, undersampling also showed promising performance for managing the global-local trade-off in time series in one of the problems.

The resampling algorithms are applied after the collection of time series is transformed into a tabular format. In effect, \texttt{TSER} is agnostic to the original structure of the time series. Although we focus on a collection of univariate time series, \texttt{TSER} is also applicable to collections of univariate time series with exogenous variables or collections of multivariate time series.

\section{Conclusions}\label{sec:6}

This work presented a novel method called \texttt{TSER} for creating synthetic time series samples. \texttt{TSER} is designed to improve the global-local trade-off in forecasting with a collection of time series with the ultimate goal of improving predictive performance on a given time series.

\texttt{TSER} intersects two relevant research topics in machine learning: forecasting with collections of time series and imbalanced domain learning.
The proposed method involves resampling the training set, composed of several time series, towards a particular time series. 

We carried out several experiments with 7 popular collections of time series.
These show that \texttt{TSER} leads to better forecasting performance relative to state-of-the-art methods. 
The main drawback of \texttt{TSER} is that, by tailoring the training set to a particular time series, the performance on other time series from the same collection is reduced. Future work will be carried out to overcome this problem. 

Overall, we believe the proposed method opens a promising research line on the intersection between global forecasting models and imbalanced domain learning.

%
%
%
\bibliographystyle{splncs04}

\end{document}